\def\year{2022}\relax
\documentclass[letterpaper]{article} 
\usepackage{aaai22}  
\usepackage{times}  
\usepackage{helvet}  
\usepackage{courier}  
\usepackage[hyphens]{url}  
\usepackage{graphicx} 
\usepackage{natbib}  
\usepackage{caption} 
\DeclareCaptionStyle{ruled}{labelfont=normalfont,labelsep=colon,strut=off} 
\usepackage{algorithm}
\usepackage{algorithmic}

\usepackage{newfloat}
\usepackage{listings}
\floatstyle{ruled}
\newfloat{listing}{tb}{lst}{}
\floatname{listing}{Listing}

\usepackage{subcaption}
\usepackage{hyphenat}
\usepackage{balance}

\usepackage{mathptmx}

\newcommand{\eg}{{e.g.}\xspace}
\newcommand{\cf}{{cf.}\xspace}

\newcommand{\vs}{{vs.}\xspace}

\newcommand*\samethanks[1][\value{footnote}]{\footnotemark[#1]}

\newcommand{\Tabref}[1]{Table~\ref{#1}}
\newcommand{\Figref}[1]{Fig.~\ref{#1}}

\newcommand{\hide}[1]{}

\newcommand{\xhdr}[1]{\vspace{1.7mm}\noindent{{\bf #1.}}}

\usepackage[show]{chato-notes}

\newcommand{\ModelName}{Homepage2Vec\xspace}

\title{Homepage2Vec: Language-Agnostic Website Embedding and Classification}
\author {
    Sylvain Lugeon\thanks{Authors contributed equally.},
    Tiziano Piccardi\samethanks{},
    Robert West  \\
}
\affiliations {
     EPFL \\
    \{sylvain.lugeon, tiziano.piccardi, robert.west\}@epfl.ch
}
\begin{document}

\maketitle
\begin{abstract}
Currently, publicly available models for website classification do not offer an embedding method and have limited support for languages beyond English. We release a dataset of more than two million category-labeled websites in 92 languages collected from Curlie, the largest multilingual human-edited Web directory. The dataset contains 14 website categories aligned across languages. Alongside it, we introduce \ModelName, a machine-learned pre-trained model for classifying and embedding websites based on their homepage in a language-agnostic way. \ModelName, thanks to its feature set (textual content, metadata tags, and visual attributes) and recent progress in natural language representation, is language-independent by design and generates embedding-based representations. We show that \ModelName correctly classifies websites with a macro-averaged F1-score of 0.90, with stable performance across low- as well as high-resource languages. Feature analysis shows that a small subset of efficiently computable features suffices to achieve high performance even with limited computational resources. 
We make publicly available the curated Curlie dataset aligned across languages, the pre-trained \ModelName model, and libraries: \url{https://github.com/epfl-dlab/homepage2vec}.

\end{abstract}

\section{Introduction}

Website classification plays an essential role for numerous purposes, from Web analysis to search engine design to Web security. However, despite the importance both for the research community and industry, the common practice is to rely on a few proprietary solutions, typically offered with a commercial license and with limited support for languages beyond English.
The goal of this paper is to release a multilingual labeled dataset collected from Curlie\footnote{\url{https://curlie.org/}} and to introduce \ModelName, a pre-trained model that supports the classification and embedding of websites starting from their homepage. Curlie is the largest human\hyp edited directory of the Web, which acts as the successor to the now-defunct DMOZ and is freely available online.
We focus our effort on collecting data and developing a language-agnostic model by selecting and combining features that do not depend on a website's language. This aspect is particularly important when dealing with Web content: \eg, a study\footnote{\url{https://w3techs.com/technologies/overview/content_language}} based on the top 10M websites estimated that in 2020 around 40\% of Web content was not in English. In addressing this challenge, recent progress in NLP and the release of pre-trained crosslingual language models enable us to revisit website classification and representation in new ways. Combined with visual properties and HTML metadata, novel NLP-based features offer an abstraction for representing Web pages without language dependency. \ModelName is based on a jointly trained neural network that returns the individual probability that a website belongs to 14 classes obtained from Curlie. Learning the individual per-class probability with a single network makes the model flexible to support at the same time binary relevance and embedding. 


With this article, we release a curated version of the Curlie dataset, the labels aligned to English, the pre-trained model, and a Python library to embed and classify any website. All the resources are available at \url{https://github.com/epfl-dlab/homepage2vec}.

\section{Related Work}
Our work is related to different topics spanning from the classification of websites to representing textual and visual features in a language-independent setup.
\label{section:rel_work}

\xhdr{Website classification} Since the very beginning of the World Wide Web's development, website classification has emerged as a hard problem. Unlike other classification tasks, where the documents come from the same source (i.e., forum messages, tweets, news feeds), websites are extremely diverse in terms of content, language, authorship, visual style, and intended audience.
Early approaches, which relied on the manual creation of taxonomies, keyword lists, and custom classification methods \cite{indyk,chekuri1997web},
were later superseded by machine learning methods, which used both textual and contextual features \cite{dumais2000hierarchical,sun_202,CombiningLinkBased,calado2006link,cai_2003,URLFeatures,NaiveBayes,shawon_2018,WithoutTheWebPage,baykan_2009}.
Other researchers proposed to extend the extraction of features beyond the properties of the current document by exploring the neighboring pages \cite{zhu_2016}. This strategy has the advantage of aiding the classifier with more contextual information when the page is minimal in content and not very informative. 
In a similar spirit to our work, researchers in the past also explored the effectiveness of visual features in classifying Web pages \cite{boer_2010}.
More recently, researchers have started to explore deep architectures based on LSTM \cite{Deng2020WebPC}, GRU \cite{du_2018}, and BERT \cite{gupta2021ensemble} applied to textual and HTML features. Although, unlike us, focusing only on a single language or a subset of very popular websites, these directions have shown how performance can be improved with the help of more complex models.

\xhdr{Multilingual text representation} While structural and visual signals can bring additional information, the textual signal remains of paramount importance in describing a website. A traditional method used in the past was to encode a document by a vector of weights proportional to the frequencies of terms (TF-IDF). 
A breakthrough in the field of textual representation was word2vec \cite{mikolov_2013}, which
offers a convenient method to obtain embedding vectors for words.
Another step forward was the development of BERT \cite{devlin2018bert}, a transformer-based neural network that has been effectively adapted to the multilingual setting (\eg, mBERT, LaBSE \cite{feng_2020}, XLM \cite{lample_2019}, XLM-R \cite{conneau2020unsupervised}). For instance, XML-R has been shown to offer language\hyp independent embeddings with performance comparable to monolingual models \cite{conneau2020unsupervised}.

\xhdr{Embedding of visual features}
Our use of visual features for website classification leverages recent advances in deep learning--based computer vision. Here, models are generally composed of a sequence of convolutional layers for feature extraction, followed by one dense layer that acts as a classifier \cite{Krizhevsky_2012,he_2015}. This scheme has the advantage that the second-to-last layer can be viewed as a low-dimensional \textit{visual embedding} of the input, which can serve as a feature for downstream tasks, such as website classification in our case.

\begin{figure}[t]
\hfill
    \begin{minipage}[t]{.49\columnwidth}
        \centering
        \includegraphics[height=3.5cm]{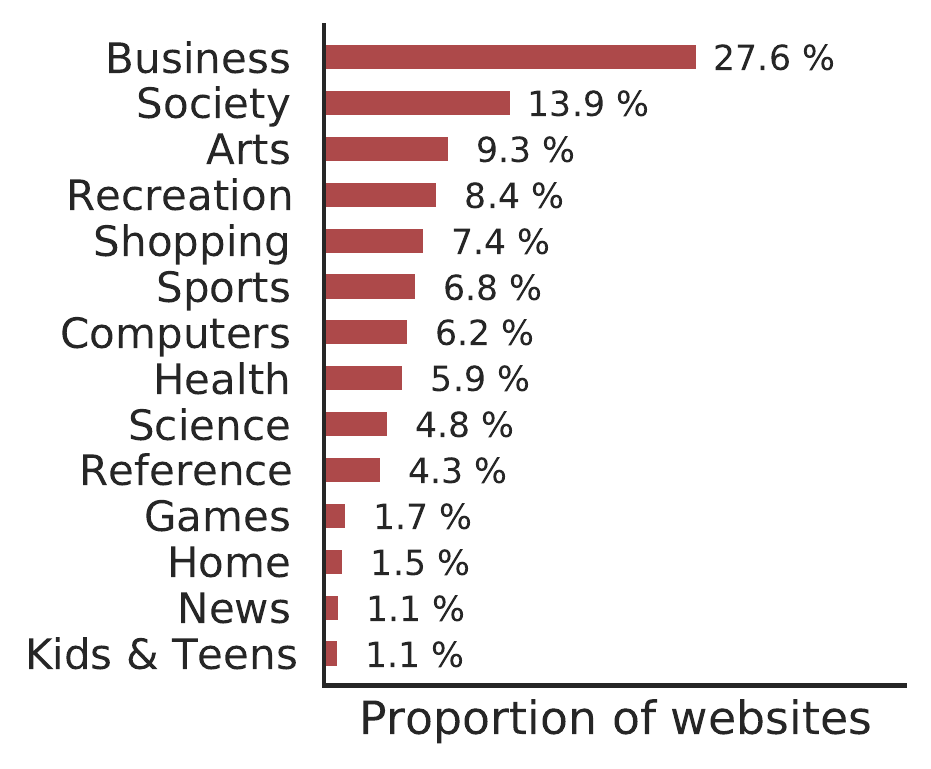}
        
        \subcaption{By category}
        \label{fig:data_distribution_category}
    \end{minipage}
    \hfill
    \begin{minipage}[t]{.49\columnwidth}
        \centering
        \includegraphics[height=3.5cm]{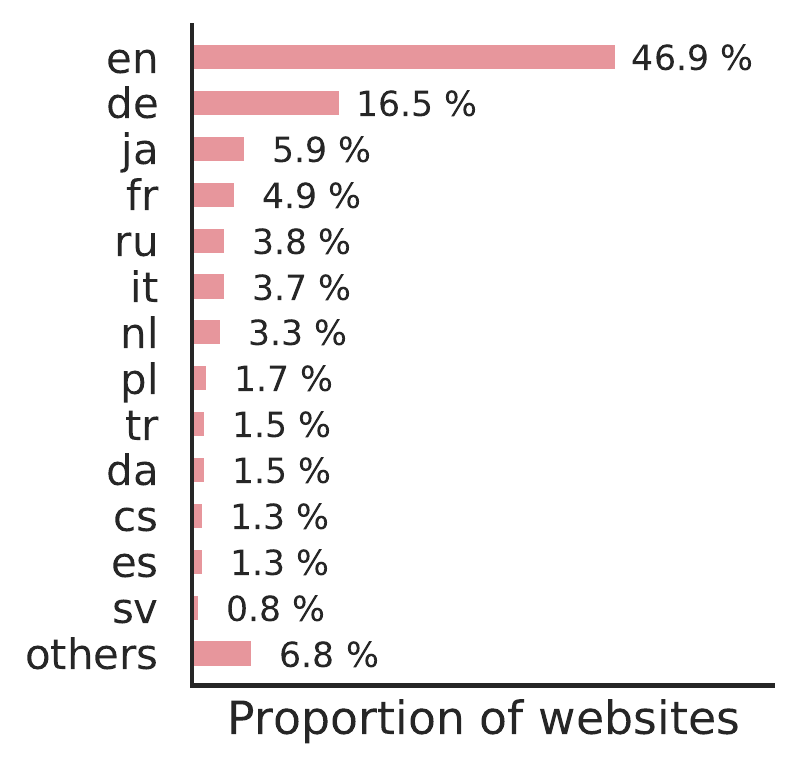}
        \subcaption{By language}
        \label{fig:data_distribution_lang}
    \end{minipage}
    \hfill
\caption{Distribution of websites in the processed Curlie dataset, with respect to (a) category and (b) language.}
\label{fig:data_distribution}
\end{figure}

\begin{figure*}[t]
    \centering
    \includegraphics[width=0.8\linewidth]{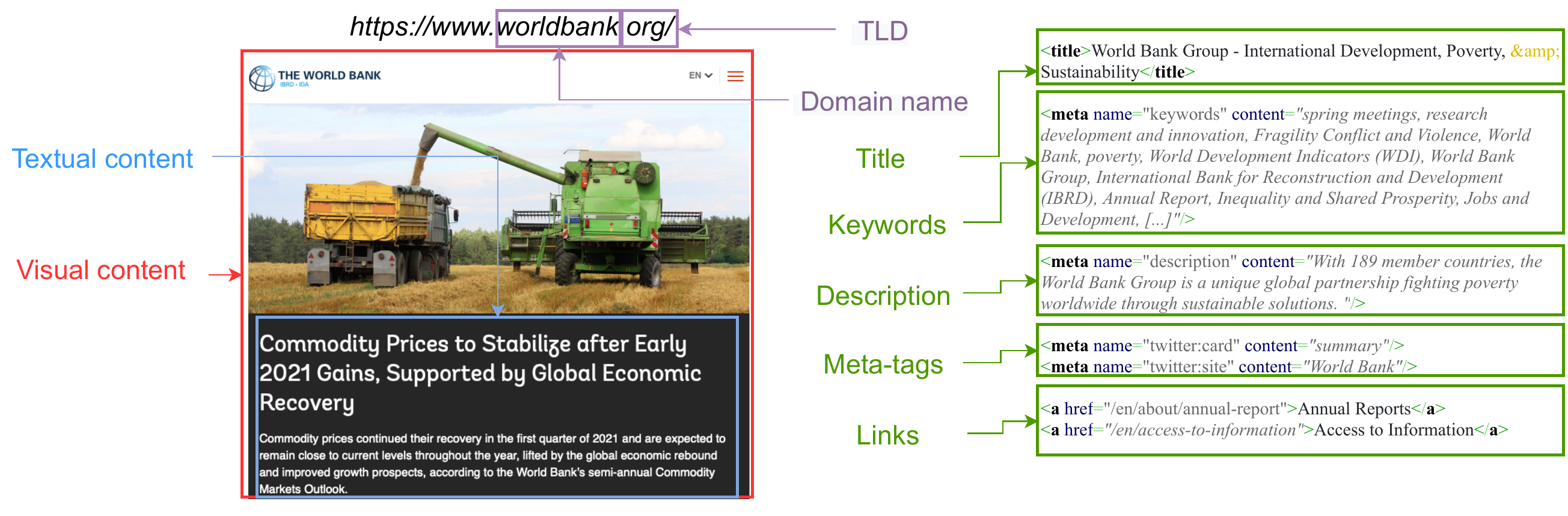}
    \caption{Features used to train \ModelName, grouped into four categories: textual features (blue), visual features (red), domain name features (purple), and Web page metadata (green).}
    \label{fig:features}
\end{figure*}

\section{Curated Curlie Dataset}
\label{Sec:Curlie}
\textit{Curlie} is an online community-edited Web directory, which serves as the successor to the now-defunct DMOZ. In total, Curlie contains more than 3M URLs in 92 languages, labeled in a hierarchical ontology of categories. The contributing community classified the websites according to 15 top-level categories: \textsc{Art}, \textsc{Business}, \textsc{Computers}, \textsc{Games}, \textsc{Health}, \textsc{Home}, \textsc{News}, \textsc{Recreation}, \textsc{References}, \textsc{Regional}, \textsc{Science}, \textsc{Shopping}, \textsc{Society}, \textsc{Sports}, and \textsc{Kids and Teens}. These categories are then organized into a hierarchical taxonomy with a high level of granularity.

We collected the Curlie directory in April 2021 by scrapping the site with Python and the support of libraries such as \textit{BeautifulSoup}.\footnote{\url{https://www.crummy.com/software/BeautifulSoup/}} We preprocessed the data to retain only unique addresses and homepages, recognized heuristically as the URLs with an empty path. This filtering leaves us 2.28M URLs, alongside their multilingual labels. Since the labels are language-specific, we obtained from the interlanguage links of Curlie the mapping between every label and its equivalent in English.
The URLs, original labels, and interlanguage mapping between labels are part of the released dataset. \Tabref{tab:global_dataset} summarizes the structure of the dataset files.

Additionally, besides the raw data, we release a processed dataset that serves as training data for \ModelName. For this data, we first drop the URLs that are non-accessible and exclude the top-level label \textsc{Regional} because its location-based nature conveys little information about the content of the website. Then, we map the label to English and assign the 14 top-level labels as categorical classes. This filtering yields 886K unique URLs. For each of these, we retrieve the HTML content and the screenshot of the page.

To generate screenshots mirroring what the user can see in the browser, we use Selenium WebDriver\footnote{\url{https://www.selenium.dev/}} to simulate the visible portion in a browser window of 1,920 $\times$ 1,080 pixels, which is standard for many computer screens. In this step, we faced the challenge of removing the modal windows typically used to display cookie information, often covering the main content. We tackle this problem with a heuristic approach by removing \texttt{div} tags whose \texttt{class} or \texttt{id} contains the substrings \texttt{popup}, \texttt{modal}, or \texttt{cookie} before capturing screenshots. 
The homepages screenshots are then processed to obtain a visual embedding based on ResNet-18 \cite{he_2015}. \Tabref{tab:processed_dataset} summarizes the schema of the files.

\begin{table}[t]
\footnotesize
\centering
\begin{tabular}{ llll }
\textbf{Files} & \textbf{Field} & \textbf{Type} & \textbf{Description} \\

\hline
Labels & url & \textit{str} & URL of the homepage\\
 & uid & \textit{int} & URL unique identifier \\
 & label & \textit{str} & Orig. Curlie full label path \\
 & lang & \textit{str} & Language of the label path \\
 \hline

Mapping & eng label & \textit{str} & Full label path in English \\
 & mapping & \textit{dict} & Mapping lang. - full path \\
 \hline
\end{tabular}
\caption{Curlie dataset (2.28M entries)}
\label{tab:global_dataset}
\end{table}

\begin{table}[t]
\footnotesize
\centering
\begin{tabular}{ llll }
\textbf{Files} & \textbf{Field} & \textbf{Type} & \textbf{Description} \\

\hline
 Content & uid & \textit{int} & URL unique identifier \\
 & html & \textit{str} & Original HTML content \\

 \hline
 Visual & uid & \textit{int} & URL unique identifier \\
 & encoding & \textit{float[]} & Visual encoding\\
 \hline
 
 Classes & uid & \textit{int} & URL unique identifier \\
 & classes vect & \textit{bool[]} & Top-level labels \\
 
\end{tabular}
\caption{Curated Curlie dataset (886K entries)}
\label{tab:processed_dataset}
\end{table}

The distribution of top-level classes and languages is shown in \Figref{fig:data_distribution_category} and \Figref{fig:data_distribution_lang}. 
The majority of the websites are associated with \textsc{Business}, \textsc{Society}, and \textsc{Arts}, while the categories with fewer instances are about \textsc{News} and \textsc{Kids and Teens}. English is the most represented language followed by German and Japanese. Although each page may, in principle, have an arbitrary number of category labels, at the top level, the data is mostly single-labeled, with only 2.1\% of samples appearing in two or more taxonomy trees of the 14 top-level classes. 





\section{\ModelName}
\label{Sec:FeaturesAndModel}

By leveraging the above-described Curlie dataset, we developed \ModelName, a multilingual model that can classify and embed any website.
This section describes the set of features used, the model architecture, and the training setup.
The Curlie dataset we release (\cf\ previous section) is enriched with the predictions made by \ModelName.

\subsection{Features}

We identify four types of features (summarized in \Figref{fig:features} and described in detail next): domain name--based, textual, visual, HTML metadata--based.
We concatenate all these features into a single high-dimensional vector to completely represent the different aspects of the website.
All features are extracted from the homepage of a website only. This is important in practice, as it means websites need not be crawled in full; only the homepage URL needs to be known and downloaded.

\xhdr{Textual content}
Webpage text is the most straightforward feature to use in website classification. We start by extracting the page's plain-text content, splitting it into sentences, and generating sentence embeddings with XLM-R \cite{conneau2020unsupervised}. Then, we compute the average of these vectors to obtain a single vector of 768 dimensions representing the document.
Since the generation of these vectors is computationally expensive, we use only the first $N=100$ sentences. We selected this threshold on a development set using the elbow method applied to the model's performance at different levels of $N$.

\xhdr{Visual features}
Visual components can play an important role in understanding the topic of a website. Humans looking at a website written in an unknown language can still often guess the type of website. Indeed, the document's layout, colors, and images can give strong hints about its content. 


To operationalize this intuition, we used the obtained images to generate a visual embedding with a pre-trained ResNet-18 model \cite{he_2015}, following the common pipeline constituted by splitting the screenshots into five crops (center plus four corners), feeding each crop to the model, and averaging the outputs. 

\xhdr{Top-level domain} 
Some top-level domains (TLD) such as \textit{.edu} or \textit{.biz} can offer a good hint about the website's content. For example, a typical use case for \textit{.edu} is university websites, whereas \textit{.biz} is commonly associated with business activities. Following this intuition, we collected from Common Crawl,\footnote{\url{https://commoncrawl.org/}} a large-scale sample of the Web, the 19 most frequent TLDs: \textit{.com, .org, .net, .info, .xyz, .club, .biz, .top, .edu, .online, .pro, .site, .vip, .icu, .buzz, .app, .asia, .gov, .space}, excluding the country code TLD (ccTLD) because they indicate geographic origin, not website content. We represent this feature with a one-hot encoding vector of 19 dimensions.

\xhdr{Domain name}
We preprocess the domain names by removing the TLD and separator symbols (``-'' and ``\_''), and splitting domain parts at the ``.''\ symbol.
When present, we drop the common ``www.''\ prefix.
After these steps, we embed the individual tokens with XLM-R and average the vectors.

\xhdr{Title, description, keywords}
Website owners are encouraged to add meta information about the content of their pages to facilitate the indexing of search engines. Since fields such as \texttt{title}, \texttt{description}, and \texttt{keywords} are typically carefully curated and very informative, we extract them from the HTML of the Web page and represent them as three feature vectors via XLM-R.

\xhdr{Links}
A website's homepage often contains links to other internal pages with explicit reference to its content. For example, by looking at the first five links of a university website, we may find \textit{EN}, \textit{About}, \textit{Education}, \textit{Research}, and \textit{Innovation}, which can be used as an indication of its academic content. We extract all the URL paths of the links on the homepage, split them into words, feed the $K=50$ most frequent words to XLM-R, and average their outputs. This approach is motivated by the fact that some parts of the link URL paths are very frequent on the page, which implicitly gives more importance to the left component of the URL path. Also, it improves the performance, as the number of links for some websites can be very large. 
Similar to the case of the textual features, we selected the best value of $K$ with the elbow method on a development set.

\xhdr{Metatags}
As in the case of description and keywords, the owners of the websites can add additional metatags to their pages to improve indexing. An example includes the tag \texttt{rating}, which is mainly associated with adult websites. We consider the 30 most frequent metatags from a sample of the Common Crawl and represent their presence with a one-hot encoding vector.



\begin{figure}[t]
    \centering
    \includegraphics[width=0.8\linewidth]{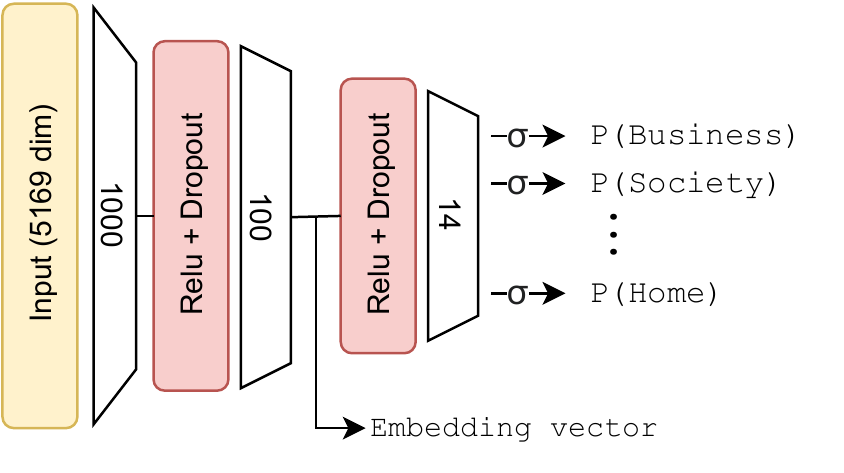}
    \caption{Neural model architecture.}
    \label{fig:architecture}
\end{figure}

\subsection{Model Architecture}

The model (\Figref{fig:architecture}) is a single neural network with an input layer of dimensionality 5,169 to accommodate all features and an output layer of size 14. The network's output represents the probability of (independently) belonging to each of the 14 categories (samples may have multiple labels). The network has two fully connected hidden layers of dimensions 1,000 and 100, respectively. We use rectified linear units (ReLU) for activation between all layers and control overfitting with a dropout probability of $0.5$.

\subsection{Training}

The training loss is defined by the average (over the 14 classes) of the binary cross-entropy. We tackle the imbalance between positive and negative samples by adapting the loss function to add a reweighting factor to the positive cases. For each positive training sample, the weight is given by the ratio between negative and positives samples in the respective class.
We train on mini-batches of size 128 using the Adam optimizer \cite{kingma2014adam} and an initial learning rate of $10^{-4}$. We decrease the learning rate by a factor of 0.1 if the loss on a held-out set of 1,000 samples decreases for 10 consecutive epochs. Finally, we stop the training when the learning rate reaches the value of $10^{-8}$.

\begin{figure}[t]
    \centering
    \includegraphics[width=\linewidth]{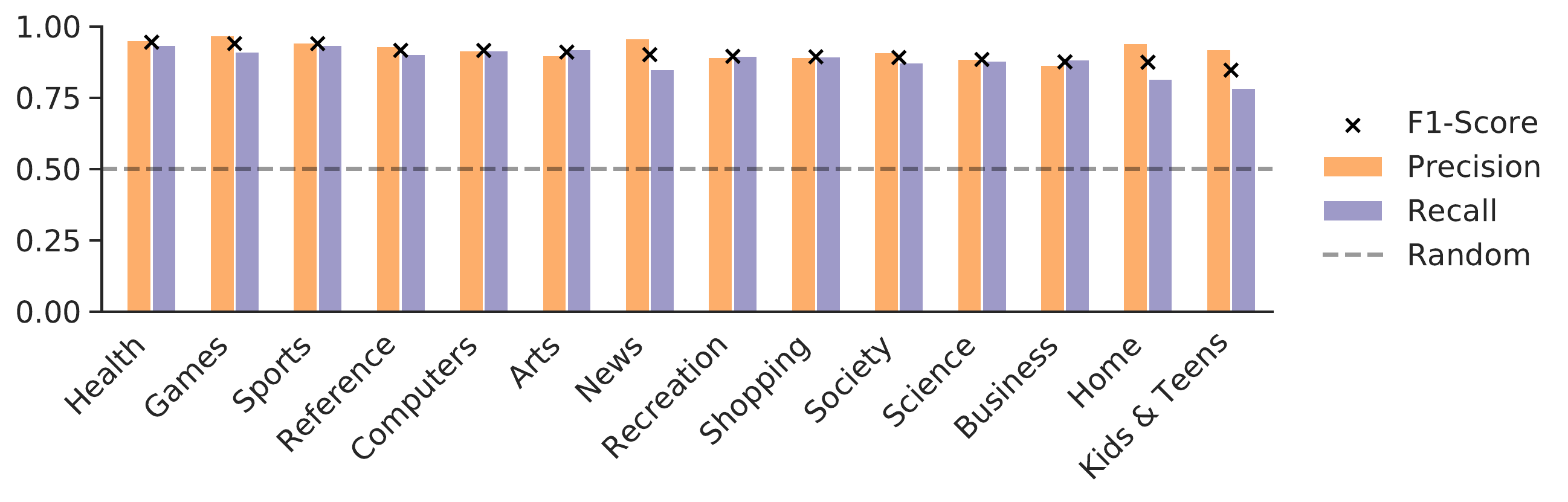}
    \caption{Performance of binary classifiers (balanced setup).}
    \label{fig:bal_eval}
\end{figure}

\begin{figure}[t]
    \centering
    \includegraphics[width=\linewidth]{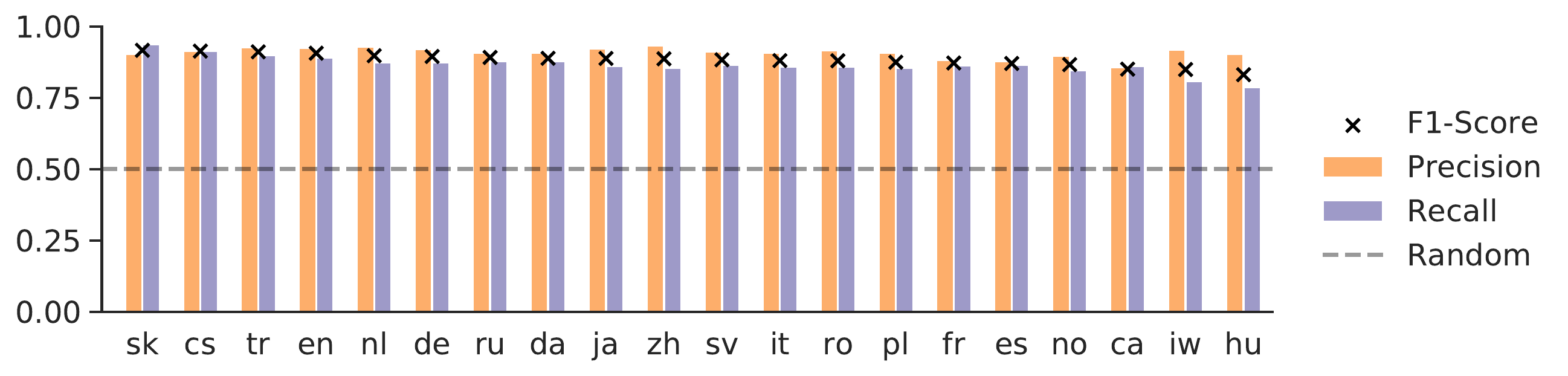}
    \caption{Macro-averaged performance by language (showing only languages with over 300 samples in the test set).}
    \label{fig:bal_eval_lang}
\end{figure}

\begin{figure*}[t]
\hfill
    \begin{minipage}[t]{.3\textwidth}
        \centering
        \includegraphics[height=3.5cm]{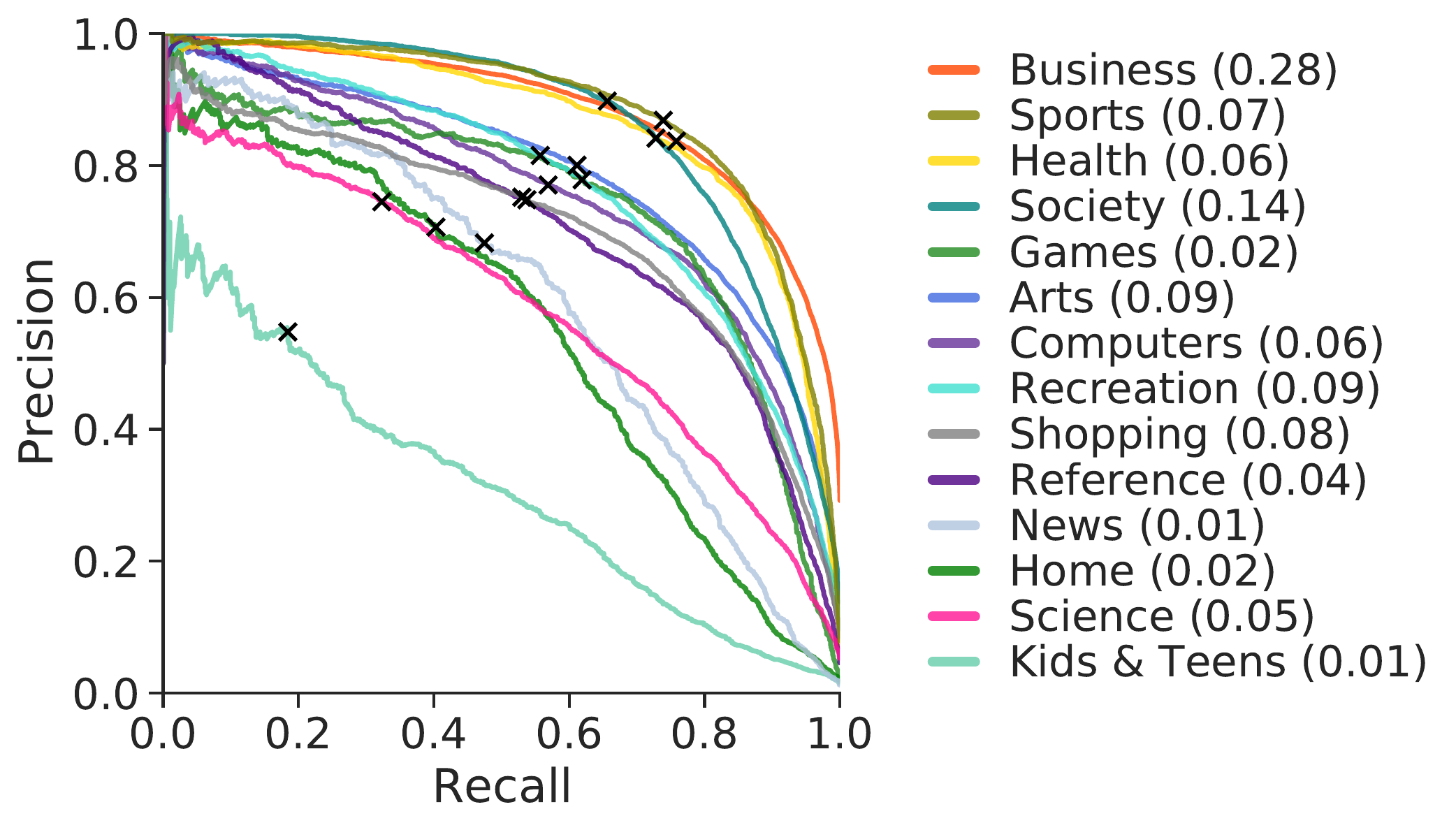}
        
        \subcaption{PR curves}
        \label{fig:pr_curves}
    \end{minipage}
    \hfill
    \begin{minipage}[t]{.69\textwidth}
        \centering
        \includegraphics[height=3.5cm]{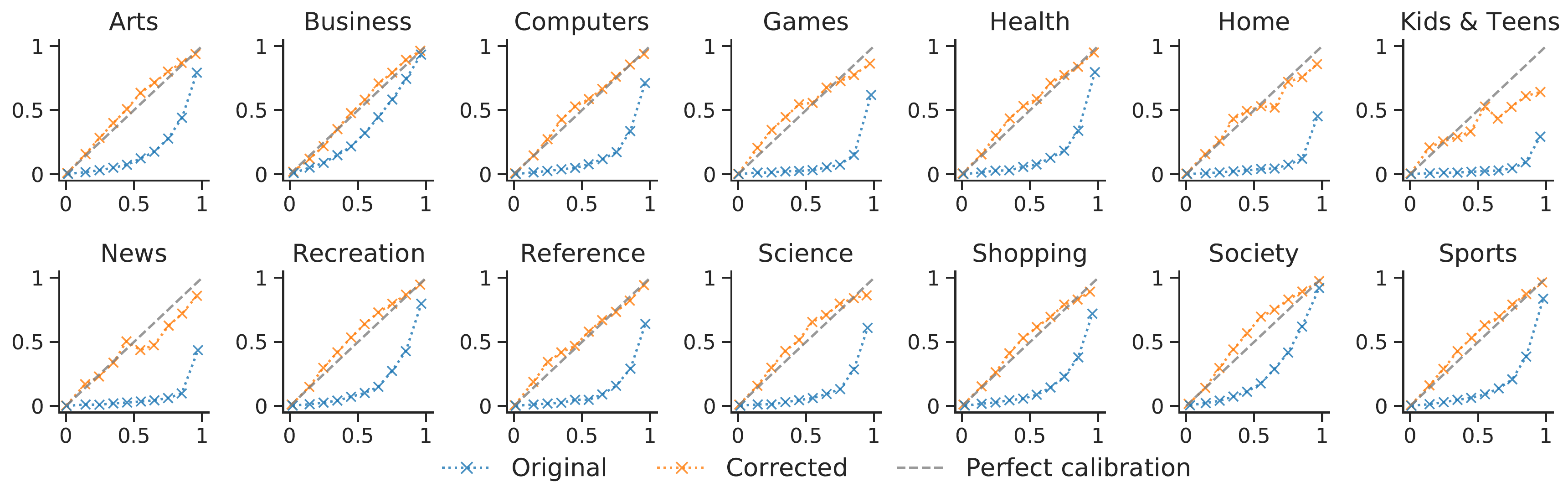}
        \subcaption{Calibration plots}
        \label{fig:calibration}
    \end{minipage}
    \hfill
\caption{(a) Precision--recall curves for evaluation with unbalanced data ($\times$: decision threshold 0.5 [after calibration]; random baseline precision [independent of recall] in parentheses). (b) Calibration plots.
}
\label{fig:performances}
\end{figure*}

\begin{figure}[t]
    \centering
    \includegraphics[width=0.6\linewidth]{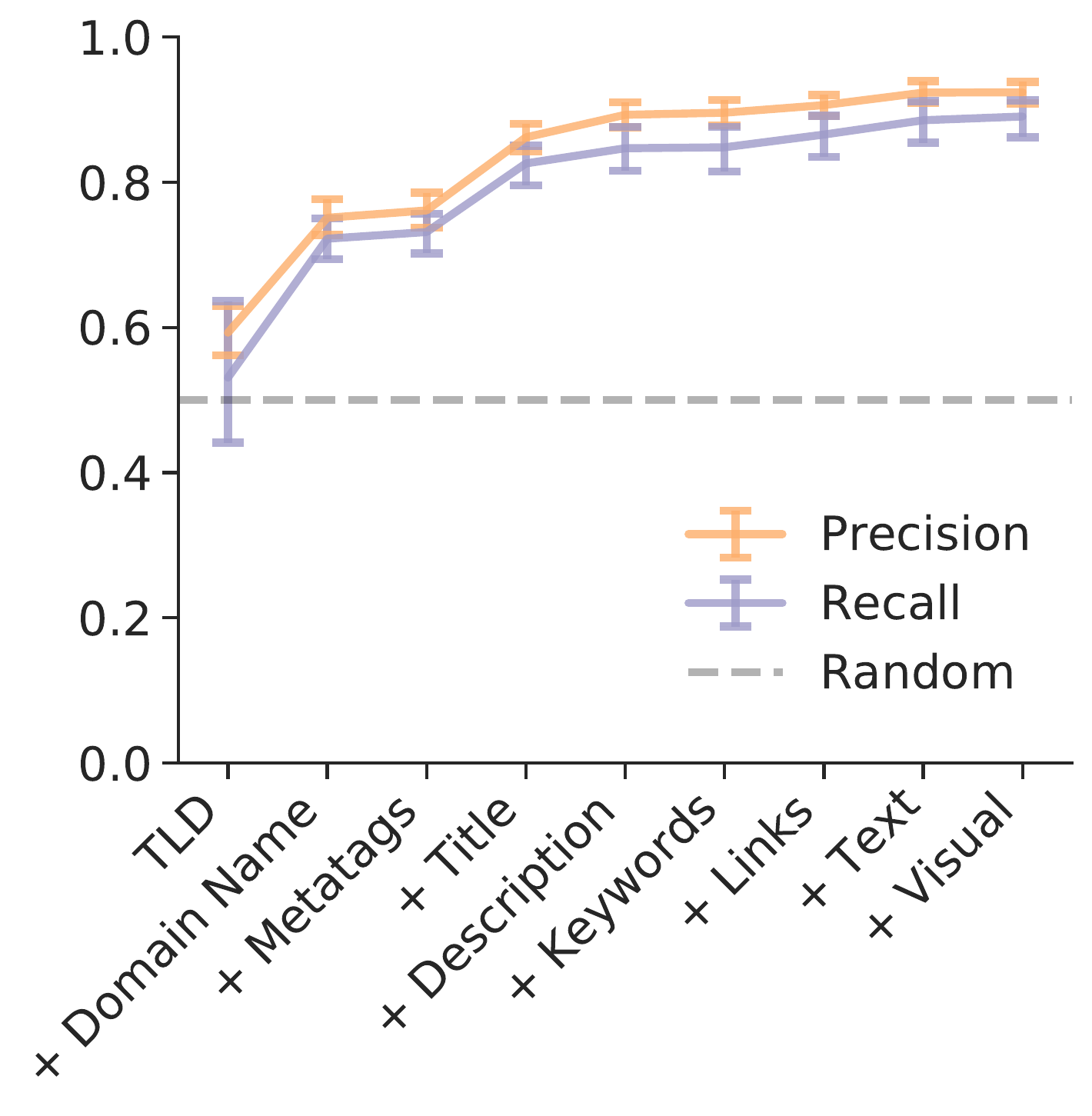}
    \caption{Macro-average metrics for incremental feature sets, with bootstrapped 95\% confidence intervals. Features are sorted left to right, from simplest to most complex.}
    \label{fig:ablation_study}
\end{figure}

\section{Evaluation}
\label{Sec:Evaluation}

\begin{figure}[t]
\hfill
    \begin{minipage}[t]{.99\columnwidth}
        \centering
        \includegraphics[width=.7\linewidth]{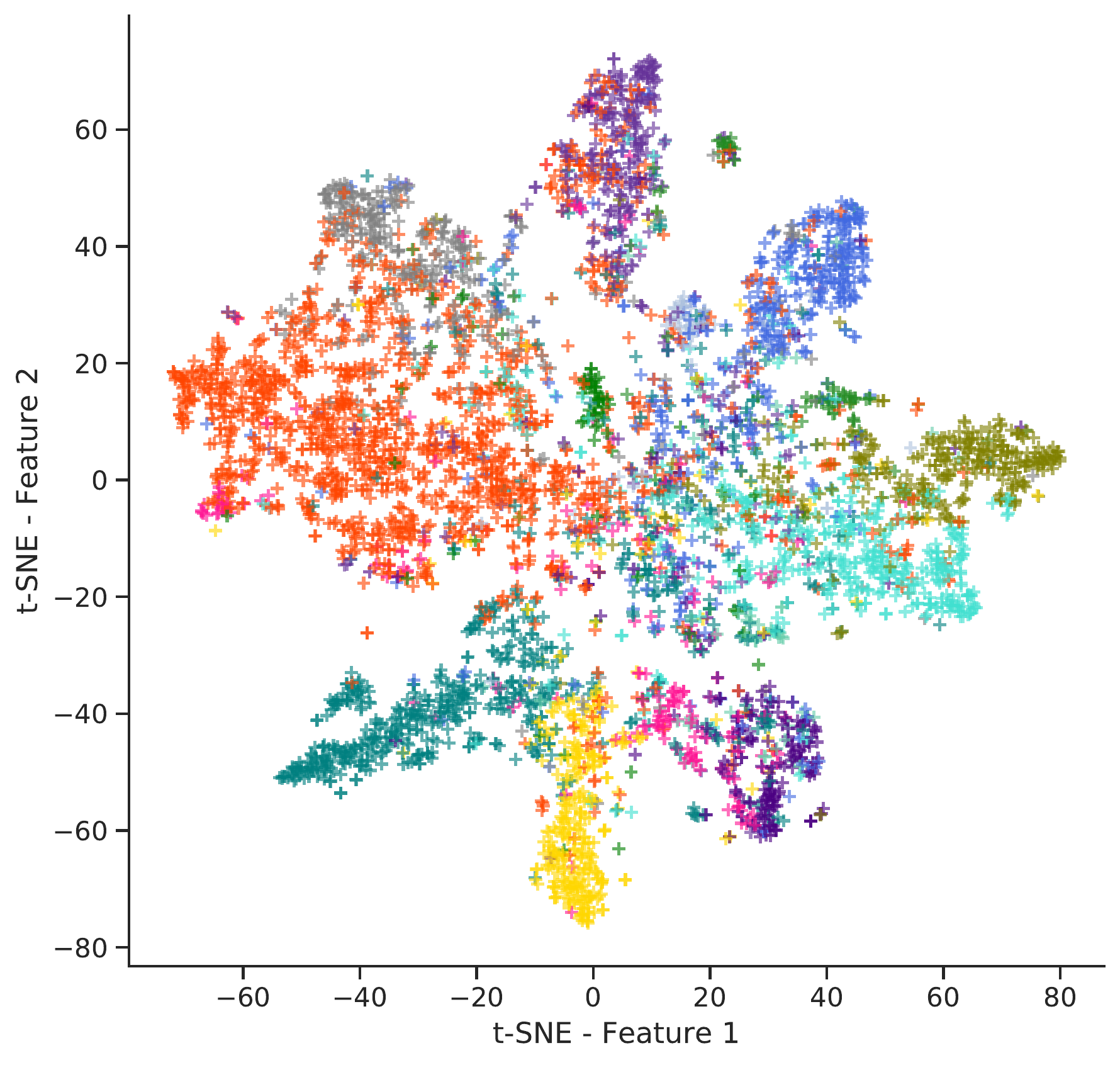}
        
        \subcaption{5k random samples}
        \label{fig:tnse_5k}
    \end{minipage}
    \hfill\\
    \begin{minipage}[t]{.99\columnwidth}
        \centering
        \includegraphics[width=.7\linewidth]{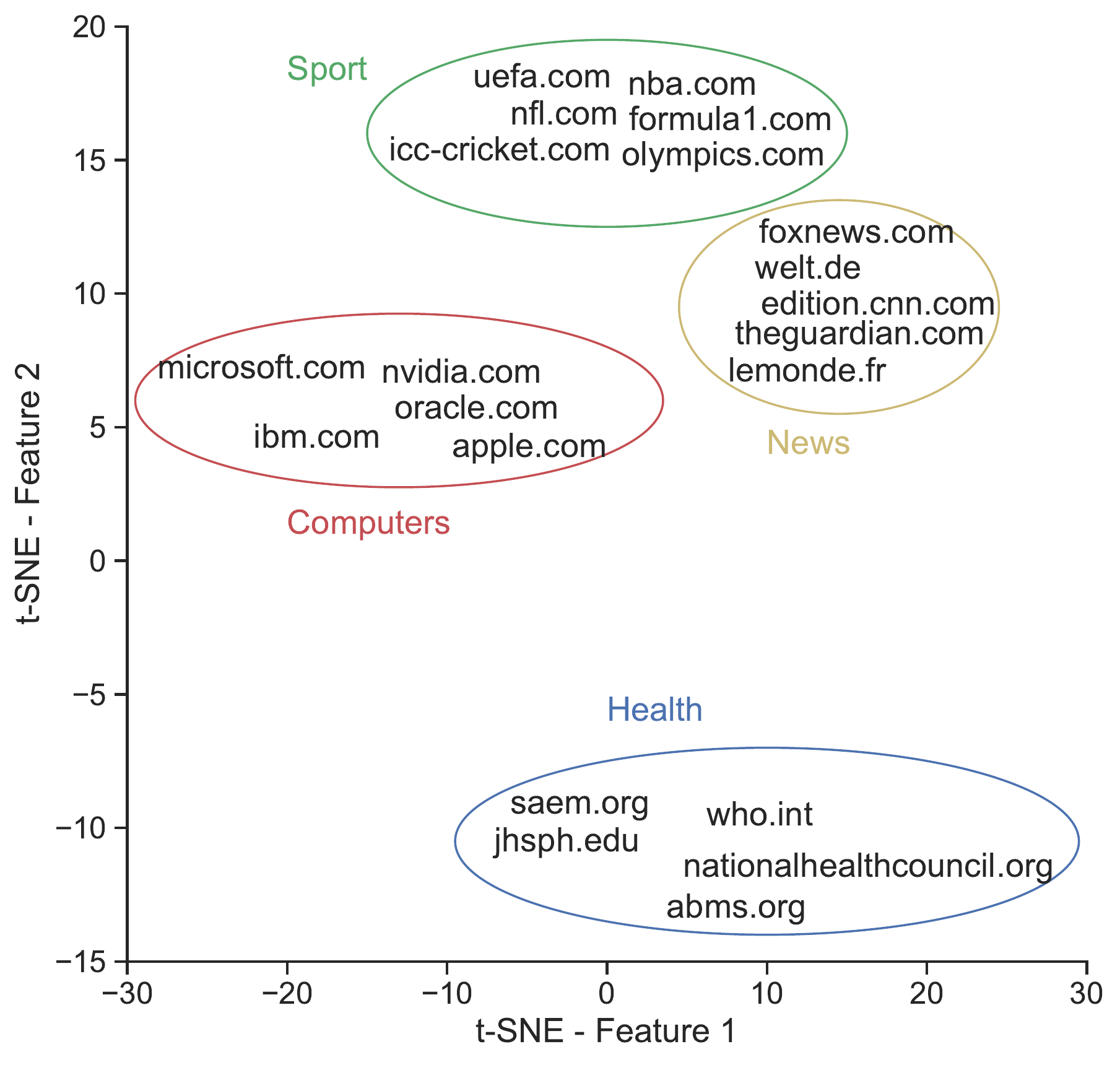}
        \subcaption{Popular domains}
        \label{fig:tsne_domains}
    \end{minipage}
    \hfill
\caption{(a) Projection in two dimensions with t-SNE of the embedding of 5K random samples from the testing set. Colors represent the 14 classes (legend in \Figref{fig:pr_curves}). (b) The projection with t-SNE of some popular websites shows that embedding vectors effectively capture website topics.}
\label{fig:EmbeddingTSNE}
\end{figure}

\subsection{Website embedding}
\label{Sec:Embedding}

Although we train and calibrate \ModelName to predict an independent probability for each class, using a single model across classes comes with the additional advantage of supporting the generation of embedding vectors. The embedding is a 100-dimensional vector, available as the output of the last hidden layer of the neural network. These vectors can be used to cluster websites, measure their distance, or study the topical distribution in a subset of the Web. \Figref{fig:EmbeddingTSNE} shows how the embedding thus obtained respects the Curlie labels and how it clusters similar websites.

\subsection{Website classification}

\xhdr{Balanced setup}
First, we evaluate the model by considering the probability returned by each category's independent binary classifier. 
We assess the individual performance in a balanced setup by balancing positive and negative samples in the testing set for each class. Performance by class is shown in \Figref{fig:bal_eval}. The macro-averaged precision is {0.920}, recall {0.886}, F1-score {0.902}, AUC\slash ROC 0.963.
Thanks to language-independent features, the model has stable performances across multiple languages, as shown in \Figref{fig:bal_eval_lang}.

\xhdr{Unbalanced setup}
Next, we evaluate the model on the entire testing set with its original, unbalanced class distribution in order to assess performance for real-world applications.

As we simulate a balanced distribution during training, the output of the classifiers must be calibrated to reflect the class priors in the testing set.  For this purpose, \citet{adjust_output_prob} proposed a simple procedure for transforming the output values of a classifier.
In the case of binary classifiers trained on a balanced distribution, the calibrated prediction for sample $i$ and class $k$ is
$\hat{s}_{ik} = s_{ik} / (s_{ik} + p_k (1 - s_{ik}))$,
where $s_{ik}$ is the original, uncalibrated prediction and $p_k$ is the ratio of negative \vs\ positive samples for class $k$ (computed on the randomly sampled training set before balancing). 
We assess calibration via the calibration plots of \Figref{fig:calibration}. The output probabilities were first grouped in bins of equal width, and then, for each bin, the fraction of positive samples is plotted against the mean value of the bin. As we see, the model is well calibrated, with calibrated curves (orange) lying close to the diagonals.
The following results (marked by $\times$ in \Figref{fig:pr_curves}) are based on a decision threshold of 0.5 on calibrated predictions.

Precision--recall curves for all classes are shown in \Figref{fig:pr_curves}. 
The results show that the evaluation of the model on unbalanced data partially reduces performance, giving a macro-average precision of 0.771, recall 0.549, F1-score 0.634, AUC\slash ROC 0.964.
The most impacted classes are the ones with only a few samples, like \textsc{Kids and Teens} (\cf\ \Figref{fig:data_distribution_category}). Two reasons primarily cause this outcome: first, discerning samples from very small classes such as \textsc{Kids and Teens} (around 1\% of the dataset) is harder than from large classes such as \textsc{Business} (27\%); and second, the testing set has missing labels. We will explore this issue and give more reassuring details in the following section.

\subsection{Feature importance}
Some of the proposed features require deep models, such as XLM-R or ResNet-18, and are computationally heavy to obtain. We want to assess the performances on the classification task if we are forced to use a limited subset of features, for example in the case of restricted resources. We first rank the features according to their complexity, where we take into account both the access (online \vs\ offline) and the computational complexity. We incrementally add the features starting from the least complex one, re-train the model, and evaluate on a balanced setup. The architecture stays fixed, only the input dimension varies to match the incremented features. The results are shown in \Figref{fig:ablation_study}. We observe that efficient performances can be achieved even without using the heaviest features. This is a useful property that can be profitable for real-world applications.

\section{Evaluation on Human Ground Truth}\label{Sec:human_eval}
Curlie, like its predecessor DMOZ, is an excellent source of labeled websites, but it is not a perfect ground truth.
Due to its human-curated nature, Curlie cannot be considered to be exhaustively labeled. Contributors add websites to a predetermined taxonomy, and in many cases, they select only one among all the relevant categories. 

Manual inspection revealed that relevant labels are frequently missing, leading to high precision but imperfect recall of the human-provided Curlie labels. 
Even the evaluation of a perfect predictive model trained and tested on the Curlie data would obtain imperfect performance when evaluated on perfectly labeled data. By manually inspecting the misclassified websites, we observed that the model tends to correctly assign a high probability to all relevant classes, even when the label is missing in the testing sample.

We further explored this observation by enriching the labels via a crowdsourced task on Amazon Mechanical Turk. We collected binary labels for all 14 categories for 807 websites.
The interface showed the website content, description, and title, and workers answered a series of binary questions. These websites originally had a total of 836 labels, according to Curlie (1.04 on average), but received 2,088 labels (2.59 on average) from crowd workers, a 2.5x increase. This lack of complete labels is consistent across categories, significantly impacting categories with few websites. For instance, while in the dataset only 11 websites were originally labeled as \textsc{Kids and Teens}, crowd workers assigned this label to 76 websites. 

The overall performance computed on the labels obtained via human annotation brings the macro-averaged precision of the unbalanced set from 0.734 to 0.873. 
These results suggest that the low precision on some classes (\Figref{fig:pr_curves}) is not so much a shortcoming of \ModelName, but rather of Curlie, and that the model can be used to enrich the original data.


\section{Library}
With the dataset and pre-trained models, we release a Python library that supports classification and embedding starting from the URL of the website. The library offers automatic content fetching, as well processing pre-fetched content, and allows users to embed it in any project with little effort. It supports classification both with and without visual features to adapt the task to the available resources. The library and usage instructions are publicly available at \url{https://github.com/epfl-dlab/homepage2vec}.

\section{Conclusion}
With this article, we release a large-scale dataset with labeled websites in 92 languages obtained from Curlie. To support the research community, we used this data to develop \ModelName, a pre-trained model for classifying and embedding websites. Due to its language\hyp agnostic features, the model performs well for low- as well as high-resource languages.
Future work to further improve the model performance includes the exploration of additional features. For example, visual features can be extended to represent the stylistic aspects of the page; \eg, websites for kids may look aesthetically very different than business sites. Similarly, metadata features could be exploited further to represent structural properties like link network and the DOM organization.
We look forward to seeing \ModelName applied in building and analyzing an increasingly polyglot World Wide Web.

\section{Ethical Statement} Curlie moderators make sure the dataset does not contain websites promoting individual products, marketing schemes, or illegal content
. Additionally, since we collected only the websites of unrestricted categories, adult content is also excluded from our dataset.

Finally, in this work, we relied on the work of human annotators on the Amazon Mechanical Turk platform. Worker compensation was in line with ethical guidelines \cite{salehi2015we,whiting2019fair}.

\balance
\bibliography{references}

\begin{thebibliography}{28}
\providecommand{\natexlab}[1]{#1}

\bibitem[{Baykan et~al.(2009)Baykan, Henzinger, Marian, and
  Weber}]{baykan_2009}
Baykan, E.; Henzinger, M.; Marian, L.; and Weber, I. 2009.
\newblock Purely URL-Based Topic Classification.
\newblock In \emph{Proc. International World Wide Web Conference (WWW)}.

\bibitem[{Cai et~al.(2003)Cai, Yu, Wen, and Ma}]{cai_2003}
Cai, D.; Yu, S.; Wen, J.-R.; and Ma, W.-Y. 2003.
\newblock Extracting Content Structure for Web Pages Based on Visual
  Representation.
\newblock In \emph{Web Technologies and Applications}, 406--417.

\bibitem[{Calado et~al.(2006)Calado, Cristo, Gon{\c{c}}alves, de~Moura,
  Ribeiro-Neto, and Ziviani}]{calado2006link}
Calado, P.; Cristo, M.; Gon{\c{c}}alves, M.~A.; de~Moura, E.~S.; Ribeiro-Neto,
  B.; and Ziviani, N. 2006.
\newblock Link-based similarity measures for the classification of Web
  documents.
\newblock \emph{Journal of the American Society for Information Science and
  Technology}, 57(2): 208--221.

\bibitem[{Calado et~al.(2003)Calado, Cristo, Moura, Ziviani, Ribeiro-Neto, and
  Gon\c{c}alves}]{CombiningLinkBased}
Calado, P.; Cristo, M.; Moura, E.; Ziviani, N.; Ribeiro-Neto, B.; and
  Gon\c{c}alves, M.~A. 2003.
\newblock Combining Link-Based and Content-Based Methods for Web Document
  Classification.
\newblock In \emph{Proc. of International Conference on Information and
  Knowledge Management (CIKM)}.

\bibitem[{Chakrabarti, Dom, and Indyk(1998)}]{indyk}
Chakrabarti, S.; Dom, B.; and Indyk, P. 1998.
\newblock Enhanced Hypertext Categorization Using Hyperlinks.
\newblock \emph{SIGMOD Rec.}, 27(2): 307–318.

\bibitem[{Chekuri et~al.(1997)Chekuri, Goldwasser, Raghavan, and
  Upfal}]{chekuri1997web}
Chekuri, C.; Goldwasser, M.~H.; Raghavan, P.; and Upfal, E. 1997.
\newblock Web search using automatic classification.
\newblock In \emph{Proc. International World Wide Web Conference (WWW)}.

\bibitem[{Conneau et~al.(2020)Conneau, Khandelwal, Goyal, Chaudhary, Wenzek,
  Guzmán, Grave, Ott, Zettlemoyer, and Stoyanov}]{conneau2020unsupervised}
Conneau, A.; Khandelwal, K.; Goyal, N.; Chaudhary, V.; Wenzek, G.; Guzmán, F.;
  Grave, E.; Ott, M.; Zettlemoyer, L.; and Stoyanov, V. 2020.
\newblock Unsupervised Cross-lingual Representation Learning at Scale.
\newblock arXiv:1911.02116.

\bibitem[{de~Boer et~al.(2010)de~Boer, van Someren, Lupascu et~al.}]{boer_2010}
de~Boer, V.; van Someren, M.; Lupascu, T.; et~al. 2010.
\newblock Classifying Web Pages with Visual Features.
\newblock In \emph{WEBIST (1)}, 245--252.

\bibitem[{Deng, Du, and Shen(2020)}]{Deng2020WebPC}
Deng, L.; Du, X.; and Shen, J. 2020.
\newblock Web page classification based on heterogeneous features and a
  combination of multiple classifiers.
\newblock \emph{Frontiers of Information Technology \& Electronic Engineering},
  21: 1004 -- 995.

\bibitem[{Devlin et~al.(2018)Devlin, Chang, Lee, and
  Toutanova}]{devlin2018bert}
Devlin, J.; Chang, M.-W.; Lee, K.; and Toutanova, K. 2018.
\newblock Bert: Pre-training of deep bidirectional transformers for language
  understanding.
\newblock \emph{arXiv preprint arXiv:1810.04805}.

\bibitem[{{Du}, {Han}, and {Zhao}(2018)}]{du_2018}
{Du}, M.; {Han}, Y.; and {Zhao}, L. 2018.
\newblock A Heuristic Approach for Website Classification with Mixed Feature
  Extractors.
\newblock In \emph{Proc of. International Conference on Parallel and
  Distributed Systems (ICPADS)}.

\bibitem[{Dumais and Chen(2000)}]{dumais2000hierarchical}
Dumais, S.; and Chen, H. 2000.
\newblock Hierarchical classification of web content.
\newblock In \emph{Conference on Research \& Development in Information
  Retrieval (SIGIR)}.

\bibitem[{Feng et~al.(2020)Feng, Yang, Cer, Arivazhagan, and Wang}]{feng_2020}
Feng, F.; Yang, Y.; Cer, D.; Arivazhagan, N.; and Wang, W. 2020.
\newblock Language-agnostic BERT Sentence Embedding.
\newblock arXiv:2007.01852.

\bibitem[{Gupta and Bhatia(2021)}]{gupta2021ensemble}
Gupta, A.; and Bhatia, R. 2021.
\newblock Ensemble approach for web page classification.
\newblock \emph{Multimedia Tools and Applications}, 1--22.

\bibitem[{He et~al.(2016)He, Zhang, Ren, and Sun}]{he_2015}
He, K.; Zhang, X.; Ren, S.; and Sun, J. 2016.
\newblock Deep residual learning for image recognition.
\newblock In \emph{Proceedings of the IEEE conference on computer vision and
  pattern recognition}, 770--778.

\bibitem[{Kan(2004)}]{WithoutTheWebPage}
Kan, M.-Y. 2004.
\newblock Web Page Classification without the Web Page.
\newblock In \emph{Proc. of International World Wide Web Conference on
  Alternate Track Papers \& Posters (WWW alt)}.

\bibitem[{Kan and Thi(2005)}]{URLFeatures}
Kan, M.-Y.; and Thi, H. O.~N. 2005.
\newblock Fast Webpage Classification Using URL Features.
\newblock In \emph{Proc. of International Conference on Information and
  Knowledge Management (CIKM)}.

\bibitem[{Kingma and Ba(2014)}]{kingma2014adam}
Kingma, D.~P.; and Ba, J. 2014.
\newblock Adam: A method for stochastic optimization.
\newblock \emph{arXiv preprint arXiv:1412.6980}.

\bibitem[{Krizhevsky, Sutskever, and Hinton(2012)}]{Krizhevsky_2012}
Krizhevsky, A.; Sutskever, I.; and Hinton, G.~E. 2012.
\newblock ImageNet Classification with Deep Convolutional Neural Networks.
\newblock In \emph{Advances in Neural Information Processing Systems},
  volume~25. Curran Associates, Inc.

\bibitem[{Lample and Conneau(2019)}]{lample_2019}
Lample, G.; and Conneau, A. 2019.
\newblock Cross-lingual language model pretraining.
\newblock \emph{arXiv preprint arXiv:1901.07291}.

\bibitem[{Mikolov et~al.(2013)Mikolov, Chen, Corrado, and Dean}]{mikolov_2013}
Mikolov, T.; Chen, K.; Corrado, G.; and Dean, J. 2013.
\newblock Efficient estimation of word representations in vector space.
\newblock \emph{arXiv preprint arXiv:1301.3781}.

\bibitem[{Rajalakshmi and Aravindan(2011)}]{NaiveBayes}
Rajalakshmi, R.; and Aravindan, C. 2011.
\newblock Naive Bayes Approach for Website Classification.
\newblock In Das, V.~V.; Thomas, G.; and Lumban~Gaol, F., eds.,
  \emph{Information Technology and Mobile Communication}, 323--326.

\bibitem[{Saerens, Latinne, and Decaestecker(2002)}]{adjust_output_prob}
Saerens, M.; Latinne, P.; and Decaestecker, C. 2002.
\newblock Adjusting the Outputs of a Classifier to New a Priori Probabilities:
  A Simple Procedure.
\newblock \emph{Neural Comput.}, 14(1): 21–41.

\bibitem[{Salehi et~al.(2015)Salehi, Irani, Bernstein, Alkhatib, Ogbe, and
  Milland}]{salehi2015we}
Salehi, N.; Irani, L.~C.; Bernstein, M.~S.; Alkhatib, A.; Ogbe, E.; and
  Milland, K. 2015.
\newblock We are dynamo: Overcoming stalling and friction in collective action
  for crowd workers.
\newblock In \emph{Proc. of Conference on human factors in computing systems
  (CHI)}.

\bibitem[{{Shawon} et~al.(2018){Shawon}, {Zuhori}, {Mahmud}, and
  {Rahman}}]{shawon_2018}
{Shawon}, A.; {Zuhori}, S.~T.; {Mahmud}, F.; and {Rahman}, M.~J. 2018.
\newblock Website Classification Using Word Based Multiple N -Gram Models and
  Random Search Oriented Feature Parameters.
\newblock In \emph{Proc of. International Conference of Computer and
  Information Technology (ICCIT)}.

\bibitem[{Sun, Lim, and Ng(2002)}]{sun_202}
Sun, A.; Lim, E.-P.; and Ng, W.-K. 2002.
\newblock Web Classification Using Support Vector Machine.
\newblock In \emph{Proc. of International Workshop on Web Information and Data
  Management (WIDM)}.

\bibitem[{Whiting, Hugh, and Bernstein(2019)}]{whiting2019fair}
Whiting, M.~E.; Hugh, G.; and Bernstein, M.~S. 2019.
\newblock Fair work: Crowd work minimum wage with one line of code.
\newblock In \emph{Proc. of Conference on Human Computation and Crowdsourcing}.

\bibitem[{Zhu et~al.(2016)Zhu, Xie, Yu, and Wong}]{zhu_2016}
Zhu, J.; Xie, Q.; Yu, S.-I.; and Wong, W.~H. 2016.
\newblock Exploiting link structure for web page genre identification.
\newblock \emph{Data Mining and Knowledge Discovery}, 30(3): 550--575.

\end{thebibliography}

\end{document}